\documentclass[10pt,twocolumn,letterpaper]{article}

\usepackage{cvpr}              

\usepackage{graphicx}
\usepackage{amsmath}
\usepackage{amssymb}
\usepackage{booktabs}

\usepackage[pagebackref,breaklinks,colorlinks]{hyperref}

\usepackage[capitalize]{cleveref}
\crefname{section}{Sec.}{Secs.}
\Crefname{section}{Section}{Sections}
\Crefname{table}{Table}{Tables}
\crefname{table}{Tab.}{Tabs.}

\def\cvprPaperID{6104} 
\def\confName{CVPR}
\def\confYear{2023}

\begin{document}

\title{DAS: Neural Architecture Search via Distinguishing Activation Score}

\author{Yuqiao Liu$^{1,3}$, Haipeng Li$^{2,3}$, Yanan Sun$^1$\thanks{Corresponding author.}, Shuaicheng Liu$^{2,3}$$^*$\\
	\normalsize$^1$ College of Computer Science, Sichuan University. \\
	\normalsize$^2$ University of Electronic Science and Technology of China. 
        \normalsize$^3$ Megvii Technology. \\
}
\maketitle

\begin{abstract}
Neural Architecture Search (NAS) is an automatic technique that can search for well-performed architectures for a specific task. Although NAS surpasses human-designed architecture in many fields, the high computational cost of architecture evaluation it requires hinders its development. 
A feasible solution is to directly evaluate some metrics in the initial stage of the architecture without any training. 
NAS without training (WOT) score is such a metric, which estimates the final trained accuracy of the architecture through the ability to distinguish different inputs in the activation layer. However, WOT score is not an atomic metric, meaning that it does not represent a fundamental indicator of the architecture. 
The contributions of this paper are in three folds. 
First, we decouple WOT into two atomic metrics which represent the distinguishing ability of the network and the number of activation units, and explore better combination rules named (Distinguishing Activation Score) DAS. We prove the correctness of decoupling theoretically and confirmed the effectiveness of the rules experimentally. Second, in order to improve the prediction accuracy of DAS to meet practical search requirements, we propose a fast training strategy. When DAS is used in combination with the fast training strategy, it yields more improvements. Third, we propose a dataset called Darts-training-bench (DTB), which fills the gap that no  training states of architecture in existing datasets. Our proposed method has 1.04$\times$ - 1.56$\times$ improvements on NAS-Bench-101, Network Design Spaces, and the proposed DTB. 
\end{abstract}

\section{Introduction}
\label{sec:intro}
The great success in deep learning in the past few years is mainly attributed to the inspiring novel network architectures designed by human experts~\cite{simonyan2014very, he2016deep}. However, this also leads to the dependence of deep learning applications on experts, hindering its development in other fields.
Neural Architecture Search (NAS)~\cite{elsken2019neural, liu2021survey} can automatically design a well-performed architecture for a specific task at hand. 
The search strategy replaces human experience, continuously selecting promising architectures and evaluating their accuracy until the best one is finally obtained.
In addition, architectures designed automatically by NAS have surpassed manual design in many domains~\cite{real2019regularized, ghiasi2019fpn}. 

\begin{figure}
    \centering
    \includegraphics[width=0.9\linewidth]{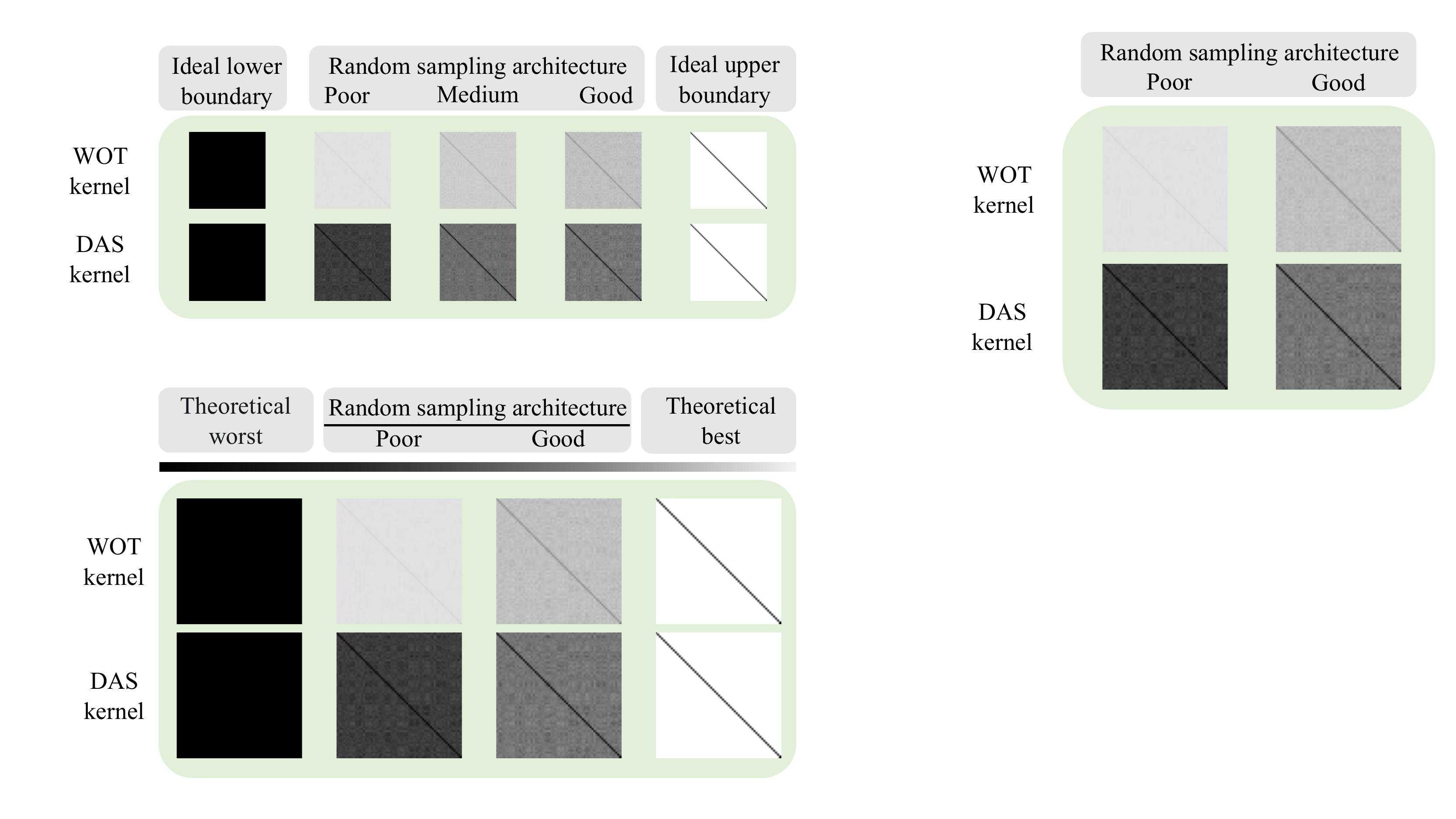}
    \caption{The comparison between kernels of WOT and DAS. Visualization of the kernel is in ascending order of final accuracy. The theoretical worst is an all-black image, meaning that the network cannot discriminate the input at all. The theoretical best image should be almost white except for the diagonal lines, implying that the architecture is powerful enough to distinguish different inputs. The kernels between the worst and best evaluate networks by sparsity, \textit{i.e.}, the whiter the visualization result corresponds to the better distinguishing ability of the network. The visualization of WOT shows that it does not accurately evaluate the architecture by sparsity, while DAS does.}
    \label{fig:teaser}
\end{figure}

Although NAS has made great achievements, its dependence on massive computing resources limits its large-scale application and research. For instance, Real \textit{et al.}~\cite{real2017large} used 250 GPUs for 11 days, and Zoph \textit{et al.}~\cite{zoph2018learning} used 450 GPUs for 4 days to complete the search. Such a large amount of computing resources are often not available. Ultimately, the reason for the large computational cost is due to inefficient evaluation methods. The most traditional and accurate method is to completely train the candidate architecture and then verify its accuracy on the validation dataset.
How to accelerate the evaluation method is a hot research topic in recent years. 
Different kinds of acceleration methods have emerged, such as low fidelity estimation~\cite{zoph2018learning,real2019regularized}, weight sharing~\cite{saxena2016convolutional,pham2018efficient}, and neural predictor~\cite{deng2017peephole,wen2020neural,Liu_2021_ICCV}.
These methods alleviate the dependence on large amounts of computation resources to some extent, but they do not get rid of the requirement to train the architecture.

Zero-cost proxies are emerging speeding-up methods. This category of method skips the computationally expensive training process, and directly evaluates the architecture in the initial state by calculating the specific statistical metric. As a result, it requires substantially lower computational cost compared to the above three methods. Given a batch of input data, after one forward (or backward) process, some information (\textit{e.g.}, gradients) in the network can be regarded as the metric. Among the zero-cost methods, WOT~\cite{mellor2021neural} performs much better on many datasets, showing great potential for development.
WOT is based on the assumption that a network in the initial state that is more discriminative of different inputs also has a higher final accuracy after training. Specifically, the discriminative ability of the network is measured by the sparsity of a symmetric kernel.
However, the WOT still has some issues to be resolved. Theoretically, it is not an atomic metric. Instead of evaluating a network by the sparsity of kernel alone, it is coupled with the number of activation units, and the visualization results are plotted in Figure~\ref{fig:teaser}. 

The main contributions of this paper are in three folds:
\begin{enumerate}
  \item We propose Distinguishing Activation Score (DAS) to evaluate searched architectures in a zero-cost manner, and further improvements can be achieved by the proposed fast training strategy.
  \item We propose a new dataset Darts-training-bench (DTB), which records the weight saving points in the early training epochs and undergoes complete training.
  \item We achieve state-of-the-art performance in multiple NAS benchmarks, including NAS-Bench-101, Network Design Spaces, and the proposed DTB.
\end{enumerate}

\section{Related Work}
\label{sec:related_work}
In this section, the development history and basic components of NAS will be introduced first. Next, the existing zero-cost proxy evaluation methods will be presented in categories.
Finally, we will present and analyze the existing benchmark datasets in NAS.

\subsection{Neural Architecture Search}
\label{subsec:nas}
Automatic neural architecture design dates back more than 20 years to the neuroevolution~\cite{floreano2008neuroevolution}, in which the optimal neural architecture and weights are searched simultaneously. The first recognized research work on NAS is the paper published by Zoph \textit{et al.}~\cite{zoph2016neural} in 2016. The main difference is that in NAS only the architecture needs to be searched, and the optimal weights are obtained by the gradient descent optimization algorithm. Thus, NAS can search for larger architectures in many real-world tasks.

In NAS, the search strategy will replace human experience by continuously sampling promising architectures from a predefined search space. The sampled architectures are evaluated for performance and then in turn guide the direction of the search strategy. In the first years of NAS emergence, more research has focused on search strategies. In chronological order, three mainstream search strategies in NAS are proposed separately, and they are reinforcement learning based~\cite{zoph2016neural}, evolutionary computation based~\cite{real2017large}, and gradient based~\cite{liu2019darts}. In recent years, research on accelerating NAS efficiency has become more popular. This is because evaluation in NAS is a time-consuming process, which hinders the large-scale practical adoption of NAS.

Traditionally, the searched architectures need to be fully trained to obtain the evaluated final accuracy. The full training process requires a large number of computational resources.
Many different kinds of speeding-up algorithms have been proposed to significantly reduce the search cost required for NAS while guaranteeing accurate evaluation.
Among the acceleration methods presented in Section~\ref{sec:intro}, the zero-cost proxy has the fastest efficiency and accurate evaluation. This is why we follow up on this category of speeding-up evaluation method.
\subsection{Zero-cost NAS}
\label{subsec:zero-cost nas}
The purpose of zero-cost NAS is to evaluate the architecture in its initial state without training. This category of the method is believed to estimate the final accuracy corresponding to the architectures based on their different reflections of the same batch of data, even in the initialized state of the neural network. According to the type of reflections or metrics used by each approach, they can be broadly divided into two categories: gradient-based metric and activation-based metric.

The gradient-based metric is inspired by parameter pruning. Snip~\cite{lee2018snip} is a saliency metric to measure the importance of parameters in a neural network in the initial state. Specifically, it approximates the saliency corresponding to a particular parameter based on the change in loss when that parameter is removed. On this basis, many gradient-based improvement approaches have been proposed~\cite{Wang2020Picking,tanaka2020pruning}. Abdelfattah \textit{et al.}~\cite{abdelfattah2020zero} utilize the saliency metrics in the evaluation strategy in NAS, summing up the saliency metrics of all parameters as a score for the architecture.

The activation-based metric has recently shown great potential and is represented by the WOT score approach~\cite{mellor2021neural}. Compared to the gradient-based metric, it is completely training-free and does not require any label or loss. It is scored based on the assumption that a batch of data is input and the network in the initial stage can divide them into different subspaces, and if the differences in data after being divided into subspaces are greater, it implies that the architecture has more potential, \textit{i.e.}, a greater chance of achieving higher final accuracy. In the WOT score, the activation state in each activation layer is calculated for each data after passing through the neural network, and then the difference between activation states is calculated as a metric to evaluate the architecture.

Since zero-cost metrics do not require a training phase, the evaluation of architectures can be done with little computational cost. It allows this category of methods to efficiently evaluate the architectures searched by different search strategies, such as random sampling or evolutionary algorithms. The search strategy used in this paper is to randomly sample a large number of architectures and finally select the one with the highest score as the architecture searched.

\subsection{Benchmark Datasets in NAS}
\label{subsec:benchmark_datasets}
NAS-Bench-101~\cite{ying2019bench} is the first publicly available benchmark dataset in the NAS research field. It designs a cell-based search space in which each architecture is a stack of cells in a prescribed skeleton. The dataset contains all possible 423K architectures in the search space, and each architecture is trained on CIFAR-10~\cite{krizhevsky2009learning} for 108 epochs to obtain final accuracy. 

Network Design Spaces (NDS) dataset~\cite{radosavovic2019network} is proposed to examine the search spaces rather than to compare search strategies like NAS-Bench-101. The dataset contains the final accuracy of several architectures in the common search spaces, such as NASNet~\cite{zoph2018learning}, AmoebaNet~\cite{real2019regularized}, PNAS~\cite{liu2018progressive}, ENAS~\cite{pham2018efficient}, DARTS~\cite{liu2019darts}. In the experiments of this paper, we make use of the data in these search spaces, and all processing is consistent with previous work~\cite{mellor2021neural}.

However, since all of the architectures in NAS-Bench-101 need to be trained, the search space is tractable and differs significantly from the common search spaces. The highest final accuracy is also much lower than that of the architecture searched by NAS methods. Though the NDS dataset contains the architectures from common search spaces, the number of training epochs is reduced and no training tricks (\textit{e.g.}, Cutout~\cite{devries2017improved}) are used, resulting in the best final accuracy in the dataset not as good as that in the original paper. The proposed dataset DTB samples architectures in the search space of DARTS, where the architectures have undergone complete training, and the final accuracy of the architectures can be comparable to the architectures searched by the state-of-the-art NAS methods.

\begin{figure*}
    \centering
    \includegraphics[width=1.0\linewidth]{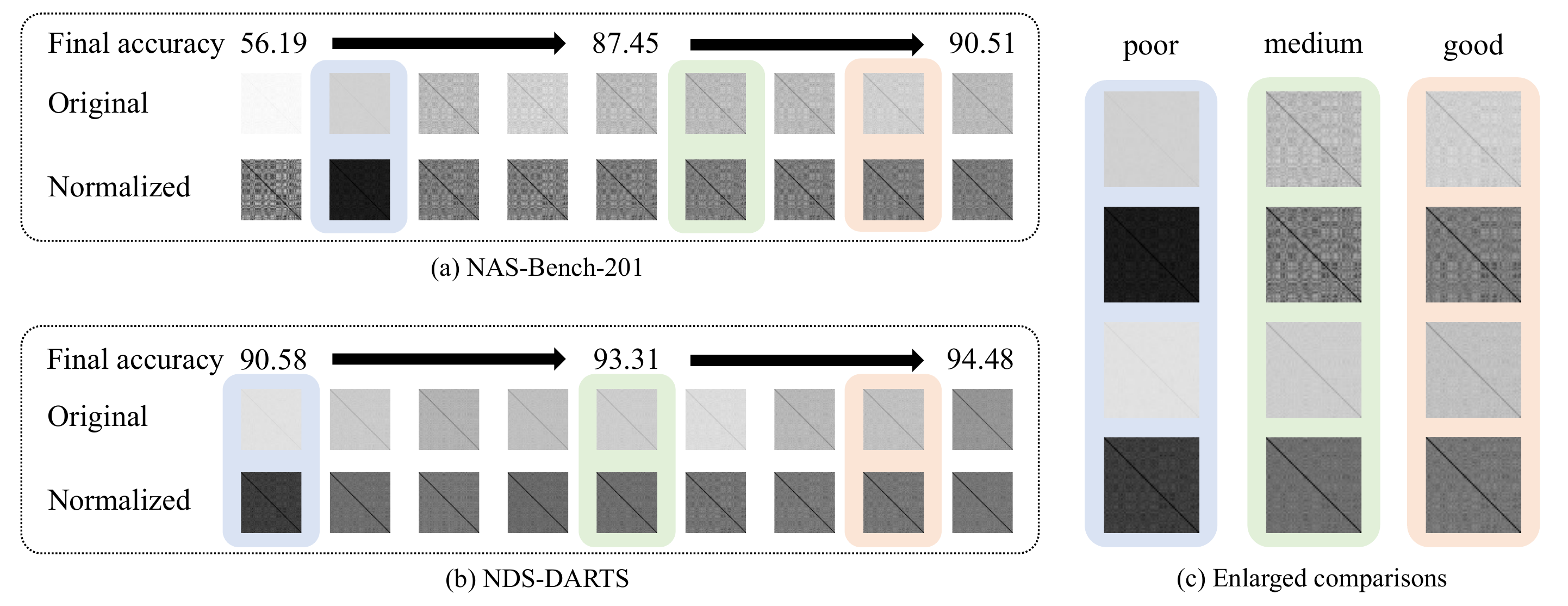}
    \vspace{-18pt}
    \caption{The visual comparison of the original and normalized kernel on two search spaces. We normalize the kernel according to the maximum value in the kernel obtained by different methods. In subfigure (a) and (b), 9 architectures are randomly selected from each search space and their corresponding kernels are shown in ascending order of final accuracy. Subfigure (c) shows the comparison of the kernel after zooming in. Theoretically, the sparser the kernel, \textit{i.e.}, the whiter the visualization, the higher the final accuracy of the corresponding architecture. The original kernel does not conform to this theory, while when normalized it satisfies.}
    \label{fig:norm_wot}
\end{figure*}

\section{Method}
\label{sec:method}
In this section, we present the motivation and formula of the WOT and propose our novel score. In order to better evaluate architectures, we theoretically decouple WOT into two comprehensive atomic metrics and find a better combination rule. This improvement does not require additional computational costs and has the potential to be combined with other atomic metrics.
Moreover, we have experimentally discovered a fast training strategy to enhance the metric, which can achieve more accurate prediction accuracy with an increased computational cost.

\subsection{WOT score}
\label{subsec:wot_score}
As introduced in Section~\ref{subsec:zero-cost nas}, WOT score evaluates the architectures by the difference in activation states of rectified linear units (ReLU) corresponding to different inputs. Each unit has two activation states: active and inactive, which we can denote by 1 and 0, respectively.
Since we can use 1 and 0 to represent the activation states of all units, when flattening these states into a one-dimensional vector in order, we can obtain a binary string. Each binary string represents the activation state of an architecture for different inputs. 
Supposing a mini-batch of data $\textbf{X}=\{x_i; {i=1, \dots,N}\}$, which passes through the network to obtain the corresponding different activation states $c_i$ encoded by the binary string, the data is divided into different linear regions.

The WOT score is based on the assumption that in the case of network initialization, if the difference between divided regions for different data is greater, the more powerful the expression of the network, and vice versa. A network with good expressive power can distinguish different inputs well even without training, and thus has a better chance of achieving higher final accuracy.

The Hamming distance $d_H(c_i, c_j)$ is used to measure the difference between activation states $c_i$ and $c_j$, because it is highly compatible with the binary string. For input data of mini-batch size $N$, the difference between all pairwise data is measured using the Hamming distance. The difference in the activation status of this batch of data is represented by kernel $K_H$, which is given by
\begin{equation}
    \footnotesize
    \label{equ:k_H}
    K_H = \left(                 
  \begin{array}{ccc}   
    N_A - d_H(c_1, c_1) & \cdots & N_A - d_H(c_1, c_N)\\  
    \vdots & \ddots & \vdots\\ 
    N_A - d_H(c_N, c_1) & \cdots & N_A - d_H(c_N, c_N)
  \end{array}
    \right) ,
\end{equation}
where $N_A$ denotes the number of activation units. The main diagonal in the kernel shows the Hamming distance $d_H(c_i,c_i)$ from itself, which is equal to zero, and after $N_A-d_H(c_i,c_i)$, the elements become $N_A$. The other elements in the kernel are also calculated in this way. When the difference between $c_i$ and $c_j$ is larger, the element value is smaller. Assuming that a kernel with fewer off-diagonal elements is defined as \textit{sparse}, then a sparser kernel indicates that the initialized network has better expressiveness.

To measure the \textit{sparsity} of the kernel, determinant $|\;.\;|$ is used to score the architecture:
\begin{equation}
    \label{equ:wot_score}
    \text{score}_\text{WOT} = \log|K_H|.
\end{equation}
When the values of the off-diagonal elements are smaller, the value of the determinant is larger, and conversely, the value of the determinant is smaller. Therefore, WOT can evaluate the architecture directly by the score.

\subsection{Decoupled WOT}
\label{subsec:de_wot}
Figure~\ref{fig:norm_wot} visualizes the kernel of randomly sampled architectures to make a comparison.
The visualization results of the kernel from two datasets are displayed in subplots (a) and (b) in Figure~\ref{fig:norm_wot}, arranged in ascending order according to the final accuracy of architectures. 
`Original' denotes the kernel used in WOT score, while `Normalized' denotes the normalized kernel in our method. In addition, we select representative kernels with different performances and enlarge them into subplot (c). It can be seen that the original kernel color is whiter in the poorly performing architectures and darker in the better-performing architectures, which does not facilitate comparison from the perspective of kernel sparsity. 
This is because the original kernel is coupled to the number of activation $N_A$, which is also the maximum value of the kernel. When we normalize the kernel to zero and one using the corresponding $N_A$, the normalized $K_H$ can be unified and comparable. After normalization, it is easier to see the difference in kernel sparsity of different performance architectures.

We note that the WOT score claims to evaluate the architecture by the sparsity of the kernel, but the calculation of the score, \textit{i.e.}, Equation (\ref{equ:wot_score}), contains an irrelevant variable, the number of activation units $N_A$. Based on the equation, we can decouple the $N_A$ theoretically:
\begin{equation}
    \label{equ:de_wot_score}
    \text{score}_\text{WOT} = \log|K_H/N_A| + N \cdot \log({N_A}).
\end{equation}
The term $K_H/N_A$ normalizes the kernel, where the element in row $i$ and column $j$ becomes $1-d_H(c_i,c_j)/N_A$, and we term it the normalized kernel $NK_H$. In addition, $d_H(c_i,c_j)/N_A$ denotes the normalized Hamming distance, which is unaffected by $N_A$ in different architectures. The second term in Equation (\ref{equ:de_wot_score}) is the decoupled $N_A$ and is multiplied by a factor, which in the original setting of WOT is equal to the size of the mini-batch $N$. It is acknowledged that the number of activation values is usually positively correlated with the final accuracy of the architecture, so it is the joint contribution of $NK_H$ and $N_A$ that gives a more accurate prediction of the WOT score.

Therefore, we naturally doubt whether the contribution coefficients of these two aspects are optimal based on the settings in the original WOT. Based on Equation (\ref{equ:de_wot_score}), we obtain the general form of decoupled WOT:
\begin{equation}
    \label{equ:de_wot_score_final}
    \text{DAS} = \log|NK_H| + \lambda \cdot \log({N_A}),
\end{equation}
where $\lambda$ is a variable coefficient. The $NK_H$ in the first term is a measure of the network's ability to discriminate between different inputs, which we refer to as distinguishing atomic. The $N_A$ in the second term is the number of activation units in the architecture, which we refer to as activation atomic. We use the combination of these two atomic metrics to score the architecture and call the scoring method \textbf{D}istinguishing \textbf{A}ctivation \textbf{S}core (DAS).

\begin{figure}
    \centering
    \includegraphics[width=1.0\linewidth]{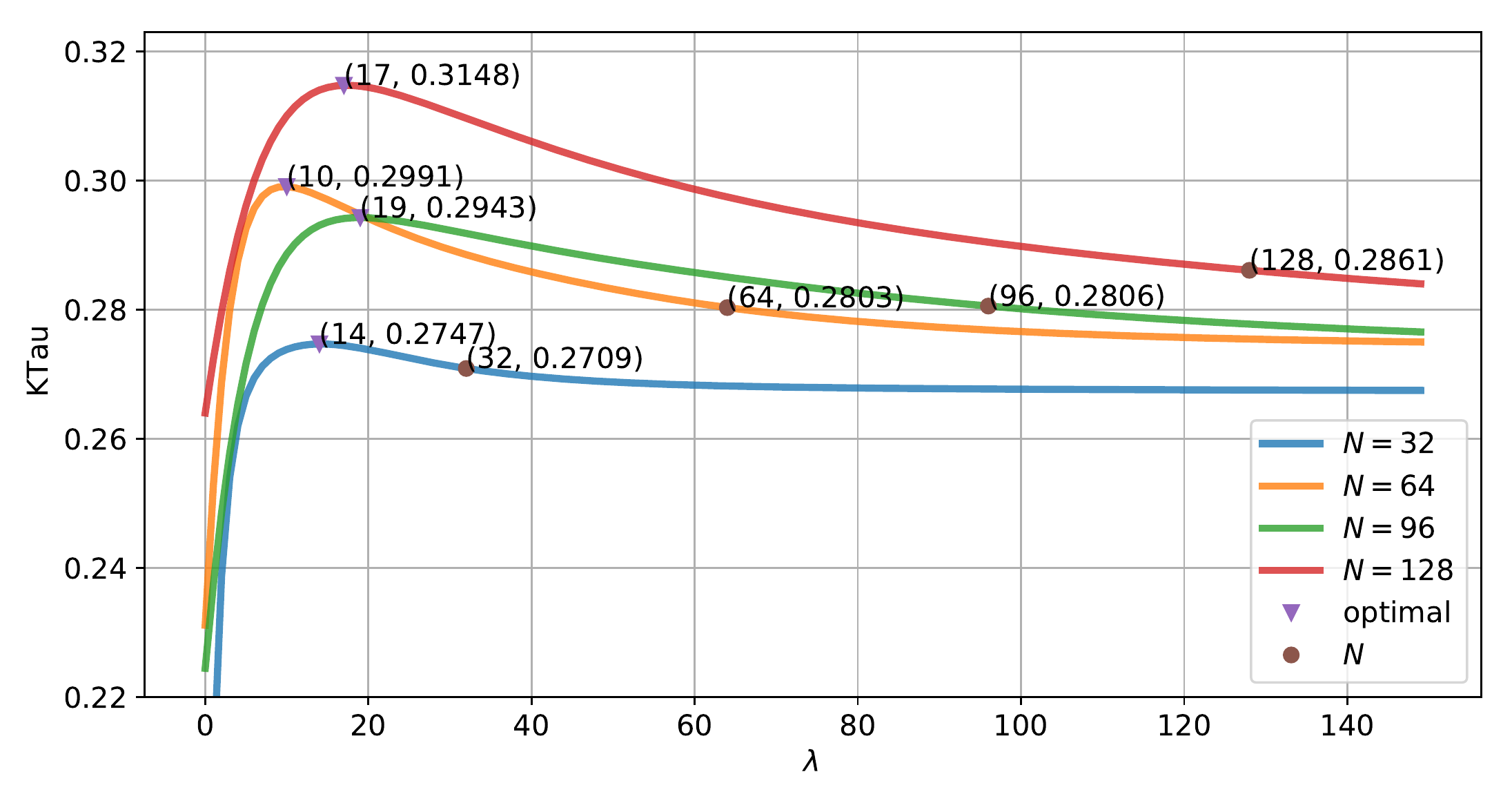}
    \vspace{-18pt}
    \caption{The relationship between optimal $\lambda$ and $N$ on NAS-Bench-101.}
    \vspace{-10pt}
    \label{fig:de_wot_batch}
\end{figure}

\begin{figure}
    \centering
    \includegraphics[width=1.0\linewidth]{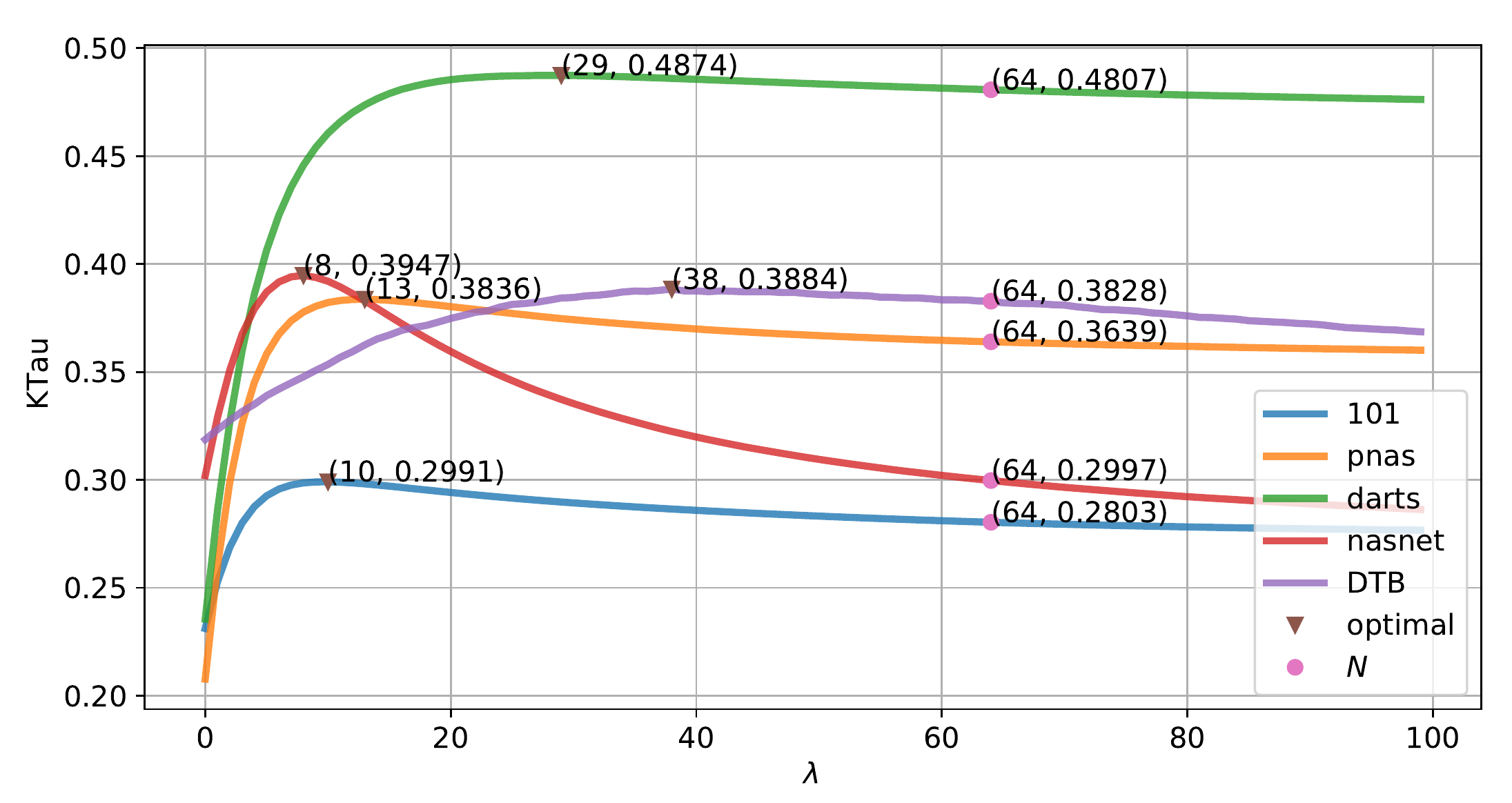}
    \vspace{-18pt}
    \caption{The grid search for the optimal $\lambda$. The $N_A$ of each search space is set as 64.}
    \label{fig:de_wot}
\end{figure}

We find in our grid-based search experiments that on the vast majority of the search spaces, the optimal $\lambda$ varies according to $N$ and is smaller than $N$. 
Two figures are shown to demonstrate this finding. Kendall’s Tau (KTau) is used to measure the evaluation accuracy of decoupled WOT, and higher KTau indicates a more accurate evaluation. 
In Figure~\ref{fig:de_wot_batch}, we show the relationship between optimal $\lambda$ and different $N$ in NAS-Bench-101. In general, the optimal $\lambda$ is proportional to $N$, \textit{i.e.}, a larger $N$ corresponds to a larger optimal $\lambda$. 
In addition, we show the search trajectory of five search spaces in Figure~\ref{fig:de_wot}. It can be found that all curves have some patterns. When $\lambda$ grows from zero, the evaluation accuracy can have a great improvement, while when $\lambda$ decreases from $N_A$, the evaluation accuracy will have a slow improvement. There will be the optimal value of $\lambda$ in the middle of zero and $N_A$.
Based on the experimental results, we empirically set the $\lambda$ to $\frac{2}{3} \cdot N$, which outperforms WOT score on most search spaces. 

\subsection{Fast Training Strategy}
\label{subsec:fast_train}
The motivation of WOT is to evaluate the architecture by the sparsity of $K_H$. It shows good capability in the initial state of the architecture, and it is reasonable to believe that after training the architecture can achieve a more accurate WOT score and thus a more precise evaluation. The experimental results in the original paper of WOT~\cite{mellor2021neural} also partially prove this theory. As the training epoch increases, the corresponding WOT score for each architecture also increases. However, the authors claim that the ranking of architectures remains similar throughout the training process, which is not consistent with the phenomenon in the figures of the original papers. Therefore, we raise a concern, about whether training could improve the evaluation accuracy.

To figure this out, we need to experiment on a dataset containing the training information, but none of the existing benchmark datasets satisfy this condition. Therefore, we propose a dataset named DTB to conduct the experiment, and the dataset is described in detail in Section~\ref{sec:dataset_DTB}. Table~\ref{tab:KTau_DTB} shows the effect of training on WOT score. We report the correlation between WOT and the final accuracy measured by KTau from the initial state (epoch=0) to the fifth epoch of training. According to the results we can see that in the first two epochs of training, there is a significant increase in the evaluation accuracy of WOT, while the accuracy in subsequent training tends to be stable.





\begin{table}
	\centering
	\caption{The performance of WOT on DTB. We calculate the WOT score for each architecture from the initial state (epoch=0) to epoch=5 and compute the evaluation accuracy measured by KTau.} 
	\label{tab:KTau_DTB}

        \begin{tabular}{c|cccccc}
            \bottomrule
            Epoch & 0 & 1 & 2 & 3 & 4 & 5    \\ \hline 
            KTau & 0.38 & 0.42 & 0.45 & 0.45 & 0.45 & 0.45  \\ \bottomrule
        \end{tabular}

\end{table}

Although training can significantly improve the evaluation accuracy of WOT, it also exponentially increases the computational cost. We count the running time of training two epochs on the DTB and scoring the initialized architectures directly\footnote{All experiments in this paper are run on an RTX 2080Ti.}. On average, the former takes 432 seconds for a single architecture, while the latter takes only 0.722 seconds, which means that training increases the computational cost by a factor of 600. Such a high computational cost is contrary to the intention of WOT.

To find a trade-off between computational cost and evaluation accuracy, we propose a fast training strategy. Only the mini-batch of data $\textbf{X}$ and the corresponding label $\textbf{Y}=\{y_i; {i=1,\cdots,N}\}$ are used to train the network for $E_f$ epochs. A network overfitted to this mini-batch of data can give a more accurate WOT score. $E_f$ is set manually and can be adjusted according to the actual situation to reduce the computational cost or improve the evaluation accuracy.

\begin{figure*}
    \centering
    \includegraphics[width=0.86\linewidth]{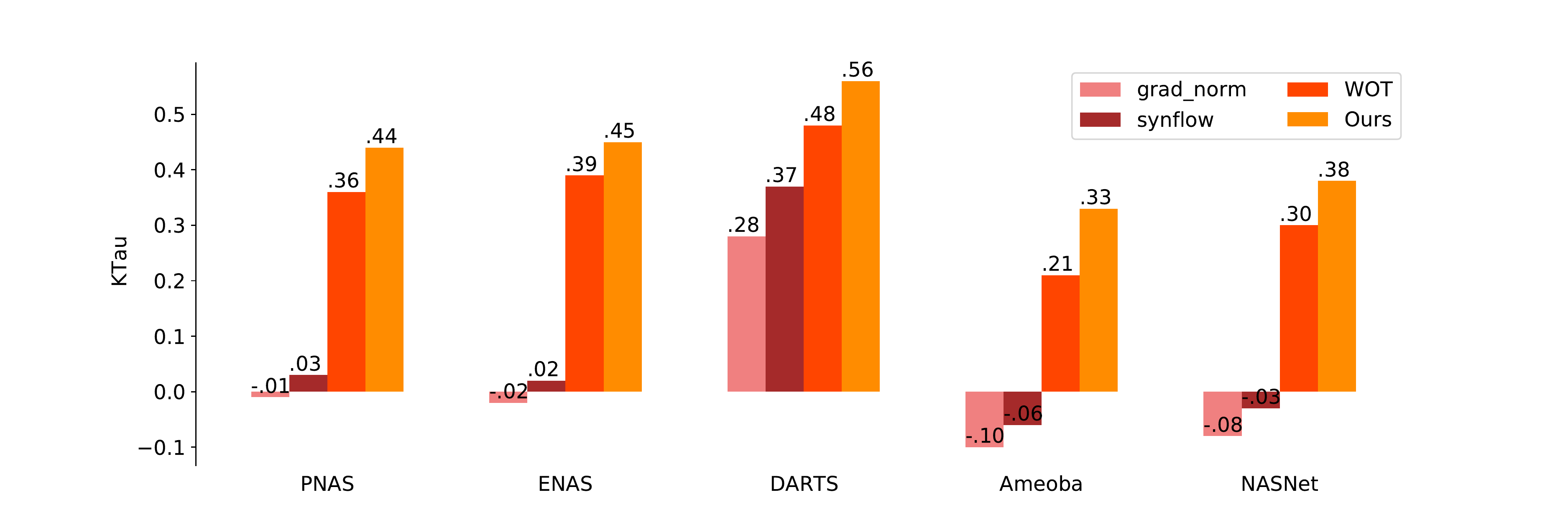}
    \caption{The comparison on five NDS search spaces. The $E_f$ is set as 30.}
    \label{fig:compare_nds}
\end{figure*}

\begin{figure*}[!ht]
\centering
     \begin{subfigure}[b]{0.23\textwidth}
         \centering
         \includegraphics[width=\textwidth]{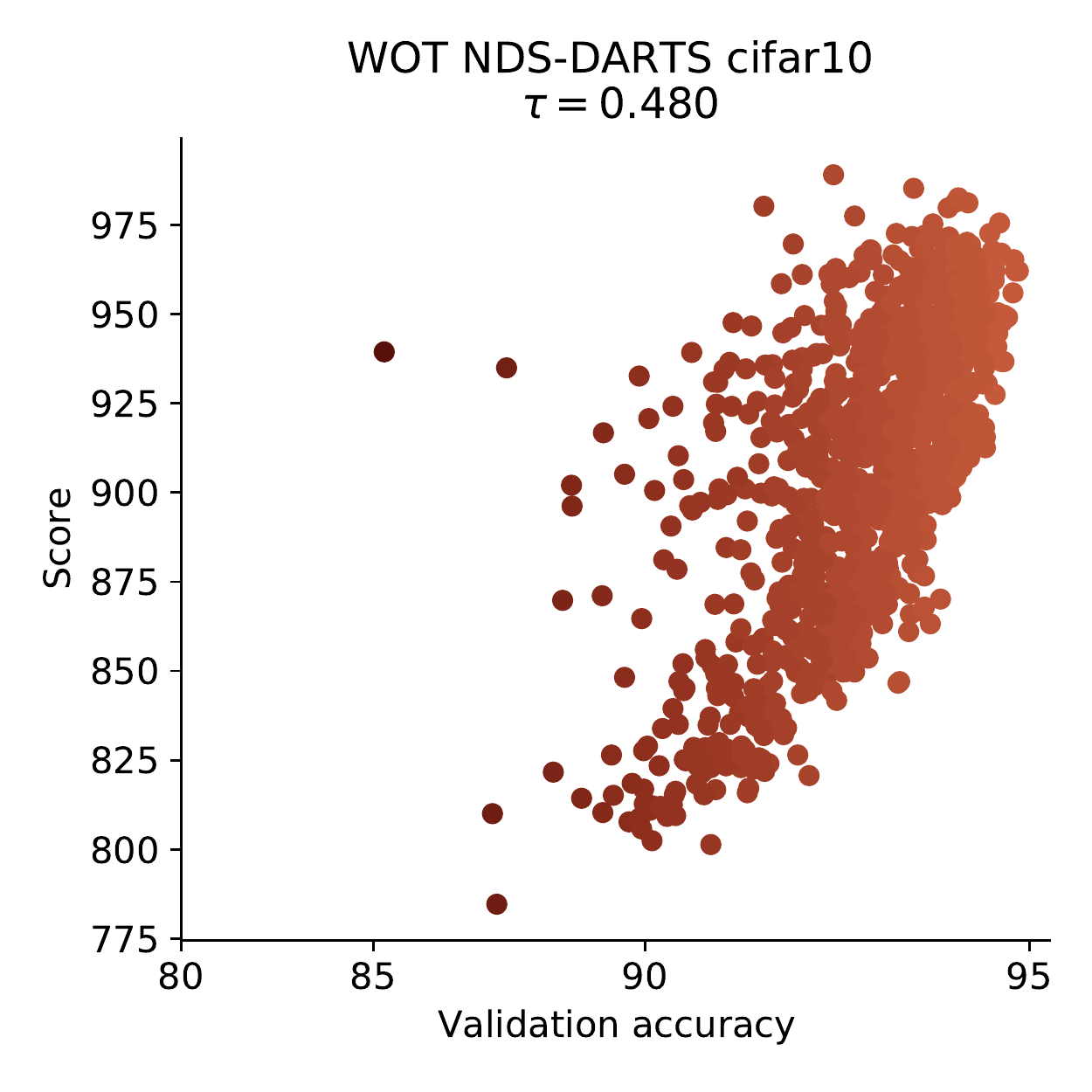}
         \caption{}
     \end{subfigure}
     \hfill
     \begin{subfigure}[b]{0.23\textwidth}
         \centering
         \includegraphics[width=\textwidth]{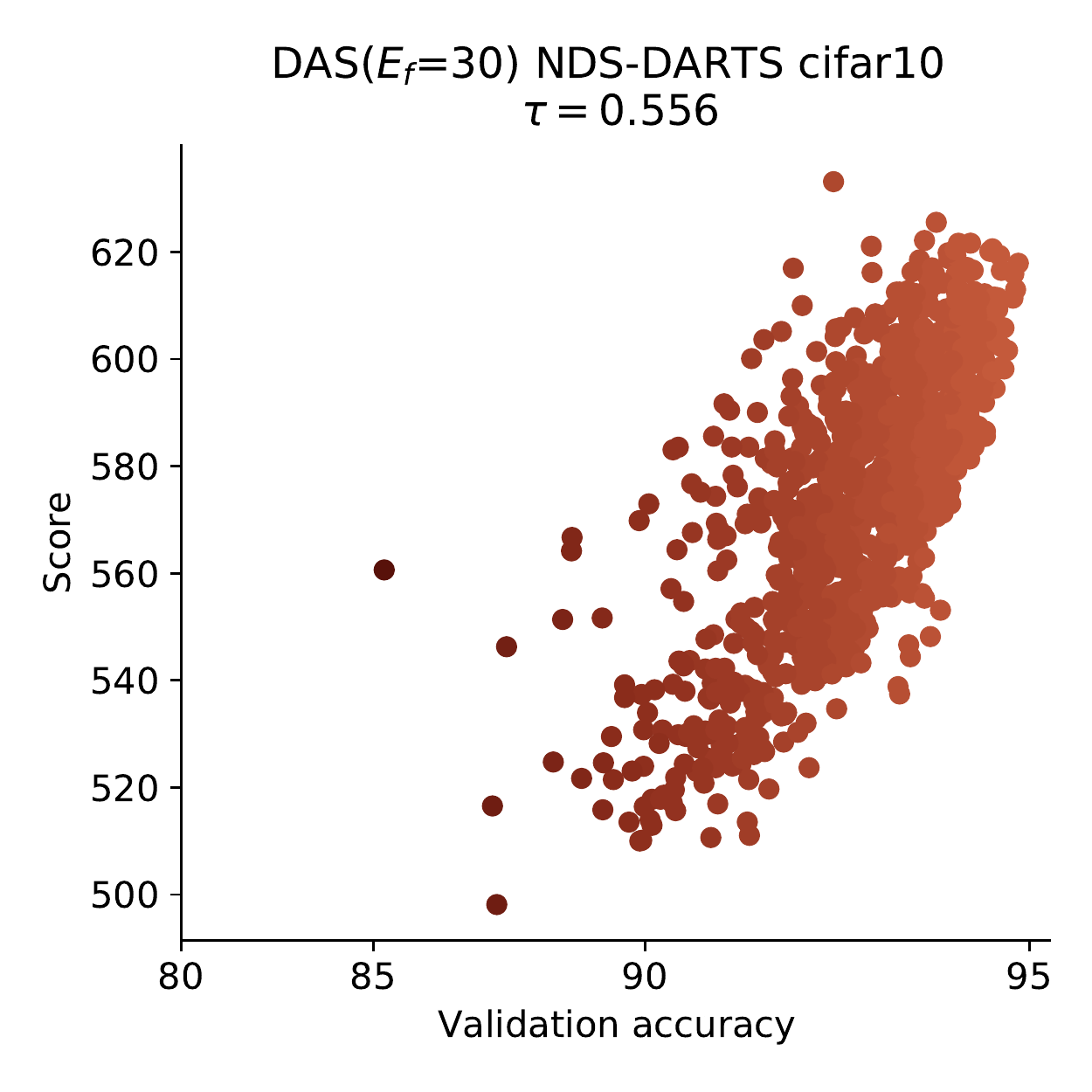}
         \caption{}
         \label{fig:darts}
     \end{subfigure}
    \hfill
     \begin{subfigure}[b]{0.23\textwidth}
         \centering
         \includegraphics[width=\textwidth]{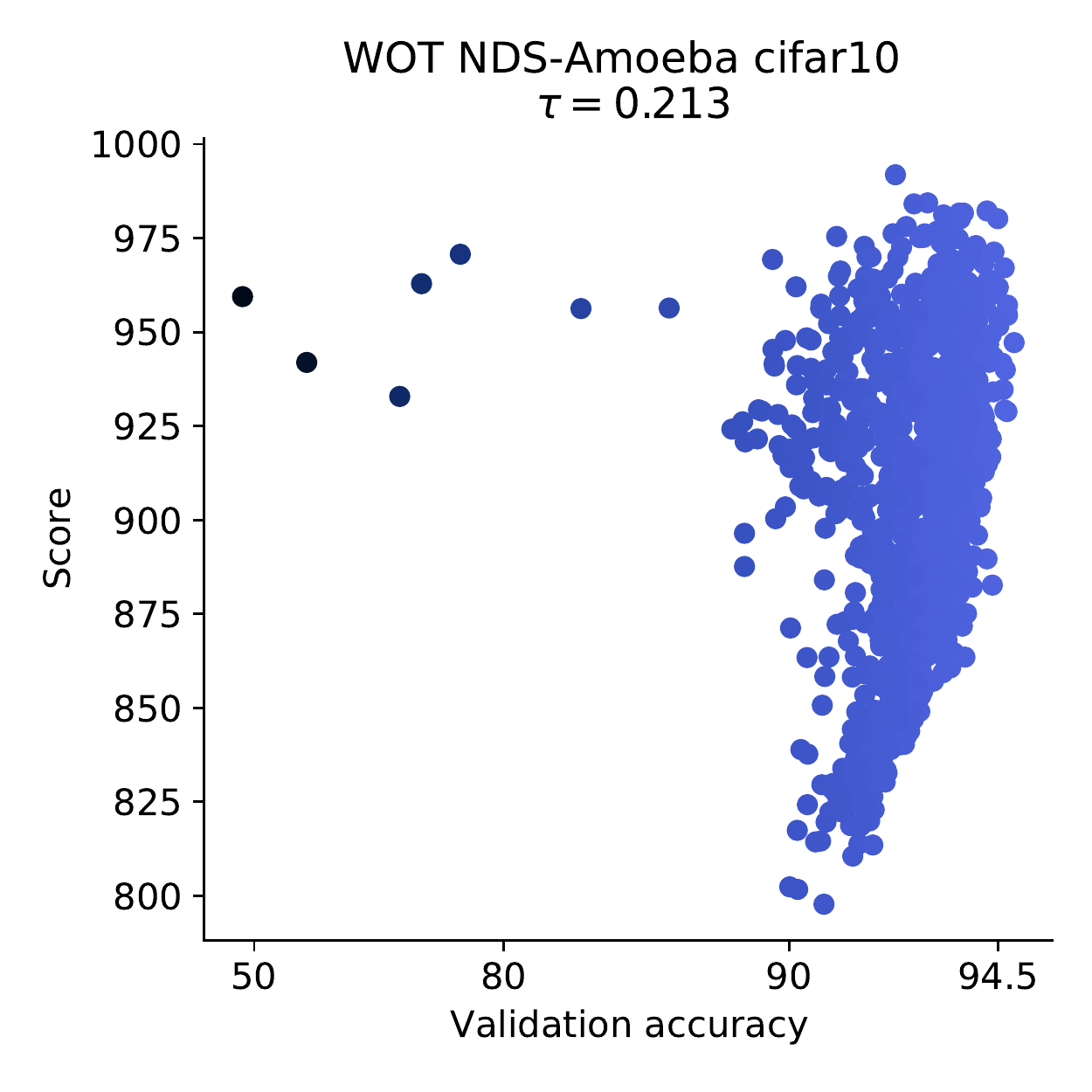}
         \caption{}
     \end{subfigure}
     \hfill
     \begin{subfigure}[b]{0.23\textwidth}
         \centering
         \includegraphics[width=\textwidth]{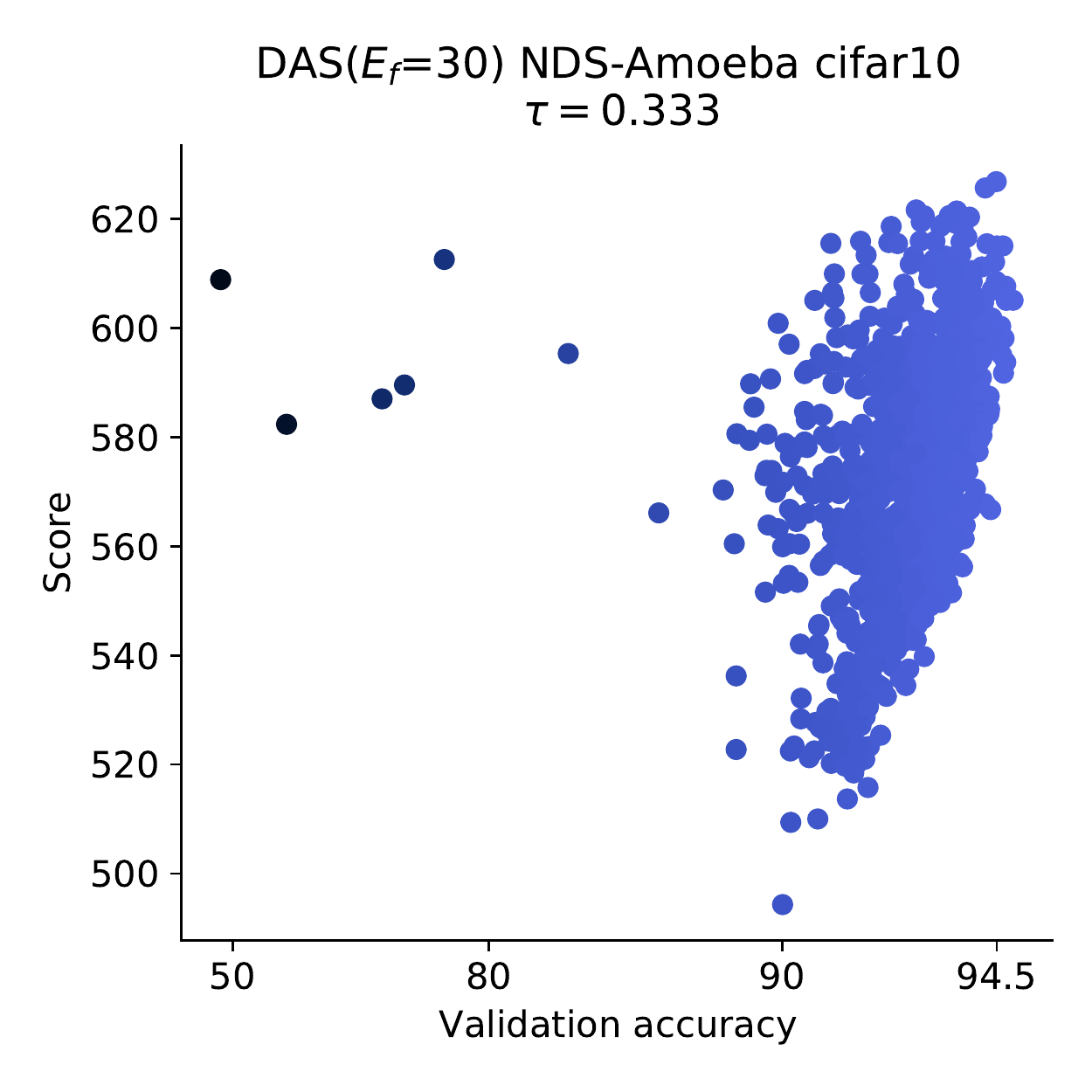}
         \caption{}
         \label{fig:amoeba}
     \end{subfigure}
     \centering
     \begin{subfigure}[b]{0.23\textwidth}
         \centering
         \includegraphics[width=\textwidth]{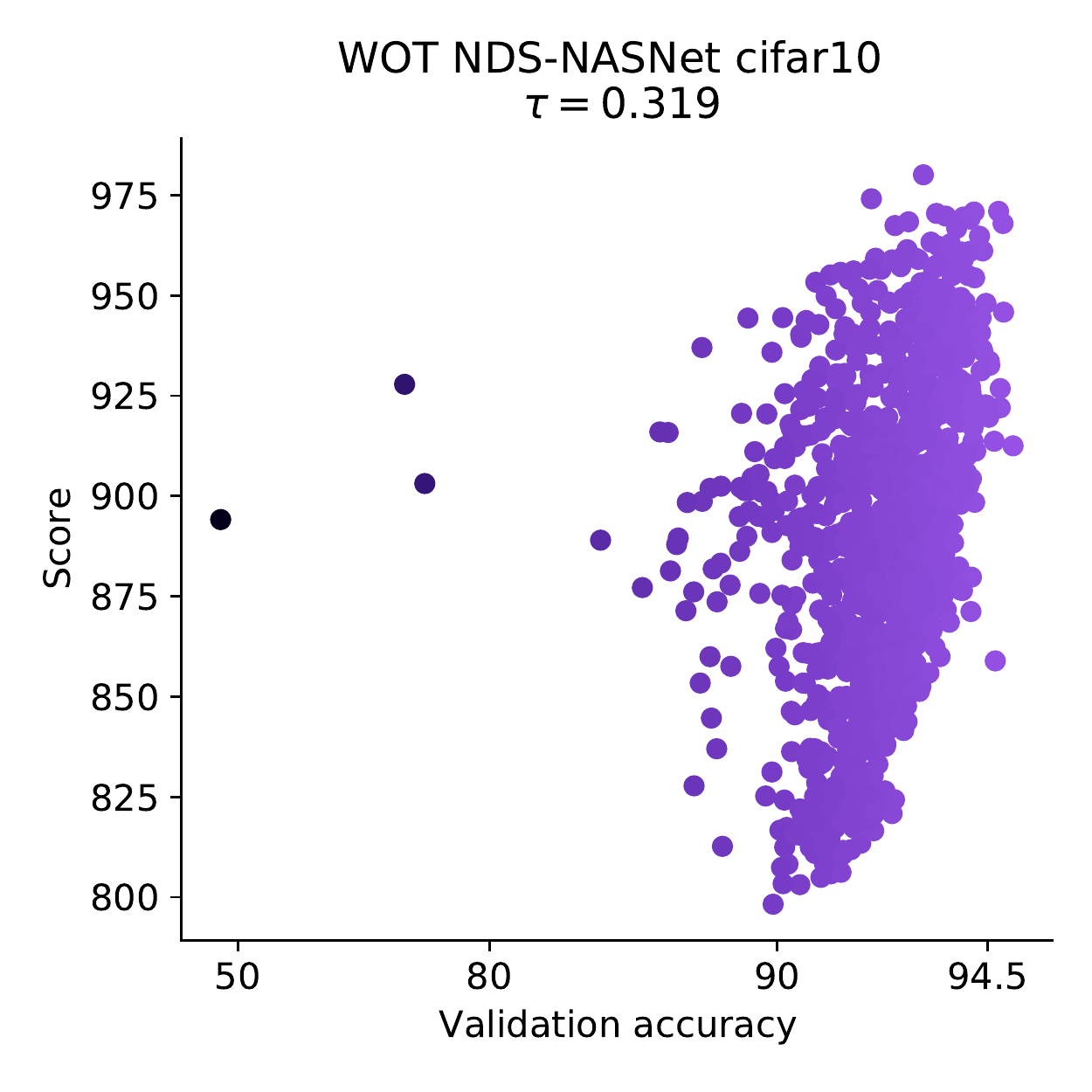}
         \caption{}
     \end{subfigure}
     \hfill
     \begin{subfigure}[b]{0.23\textwidth}
         \centering
         \includegraphics[width=\textwidth]{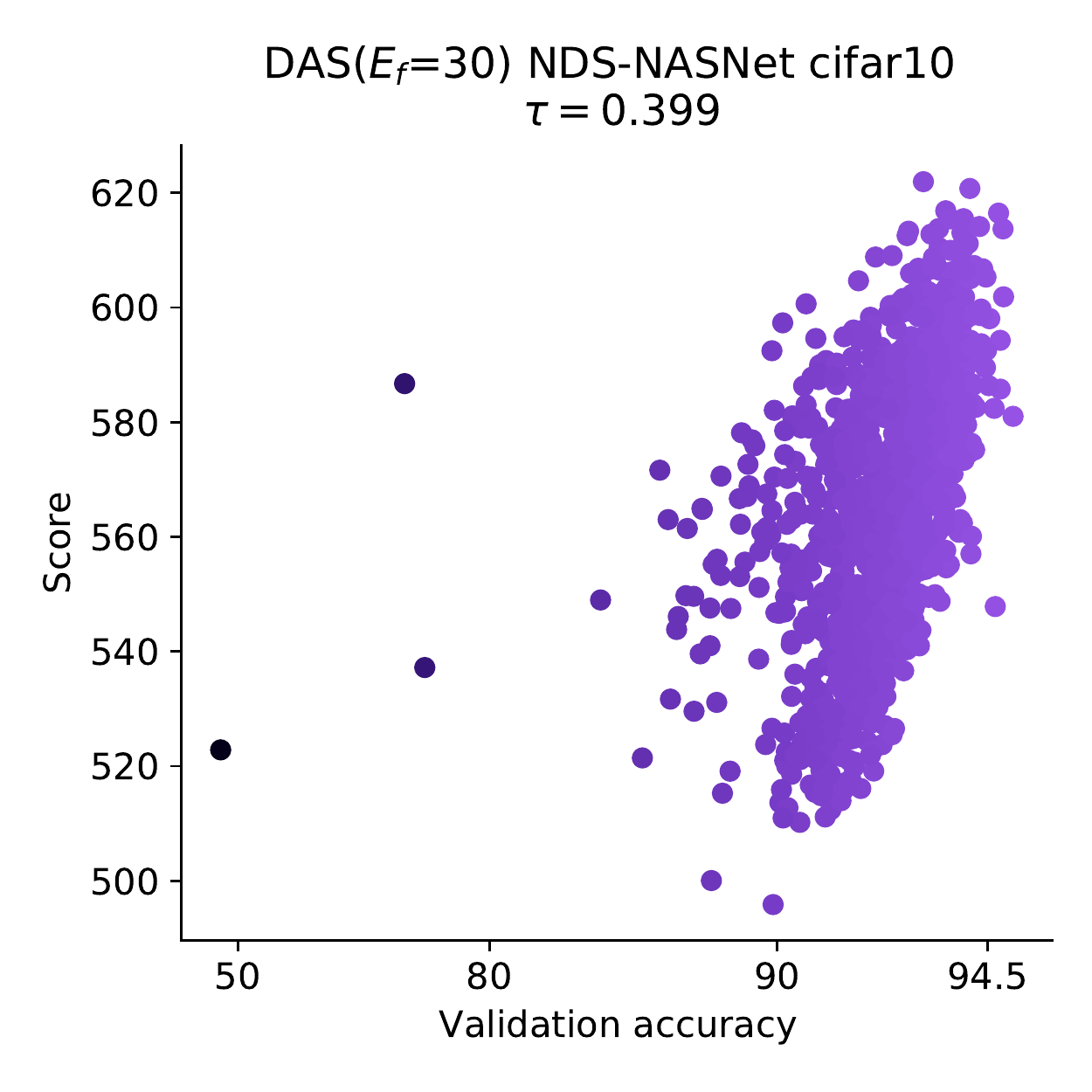}
         \caption{}
         \label{fig:nasnet}
     \end{subfigure}
    \hfill
     \centering
     \begin{subfigure}[b]{0.23\textwidth}
         \centering
         \includegraphics[width=\textwidth]{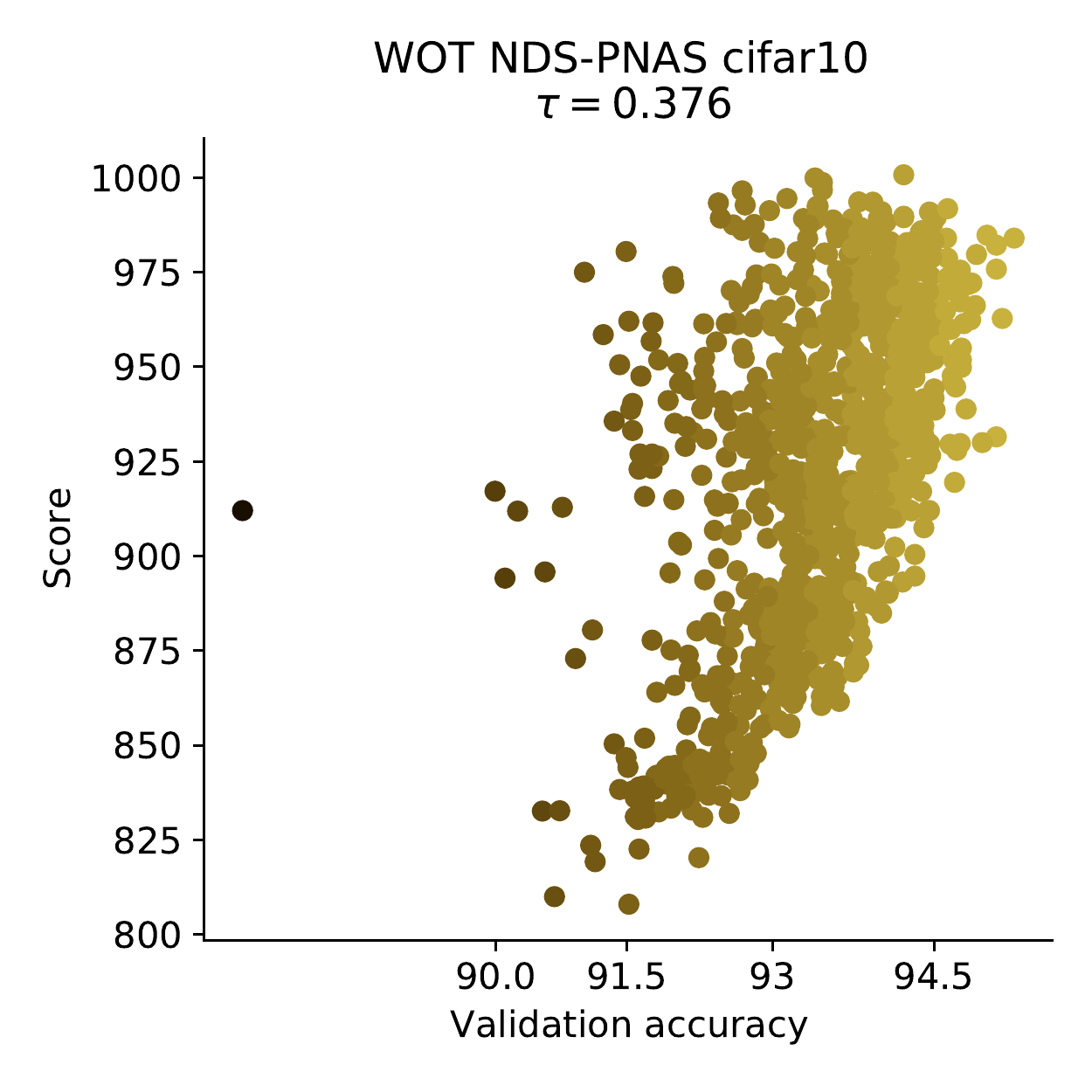}
         \caption{}
     \end{subfigure}
     \hfill
     \begin{subfigure}[b]{0.23\textwidth}
         \centering
         \includegraphics[width=\textwidth]{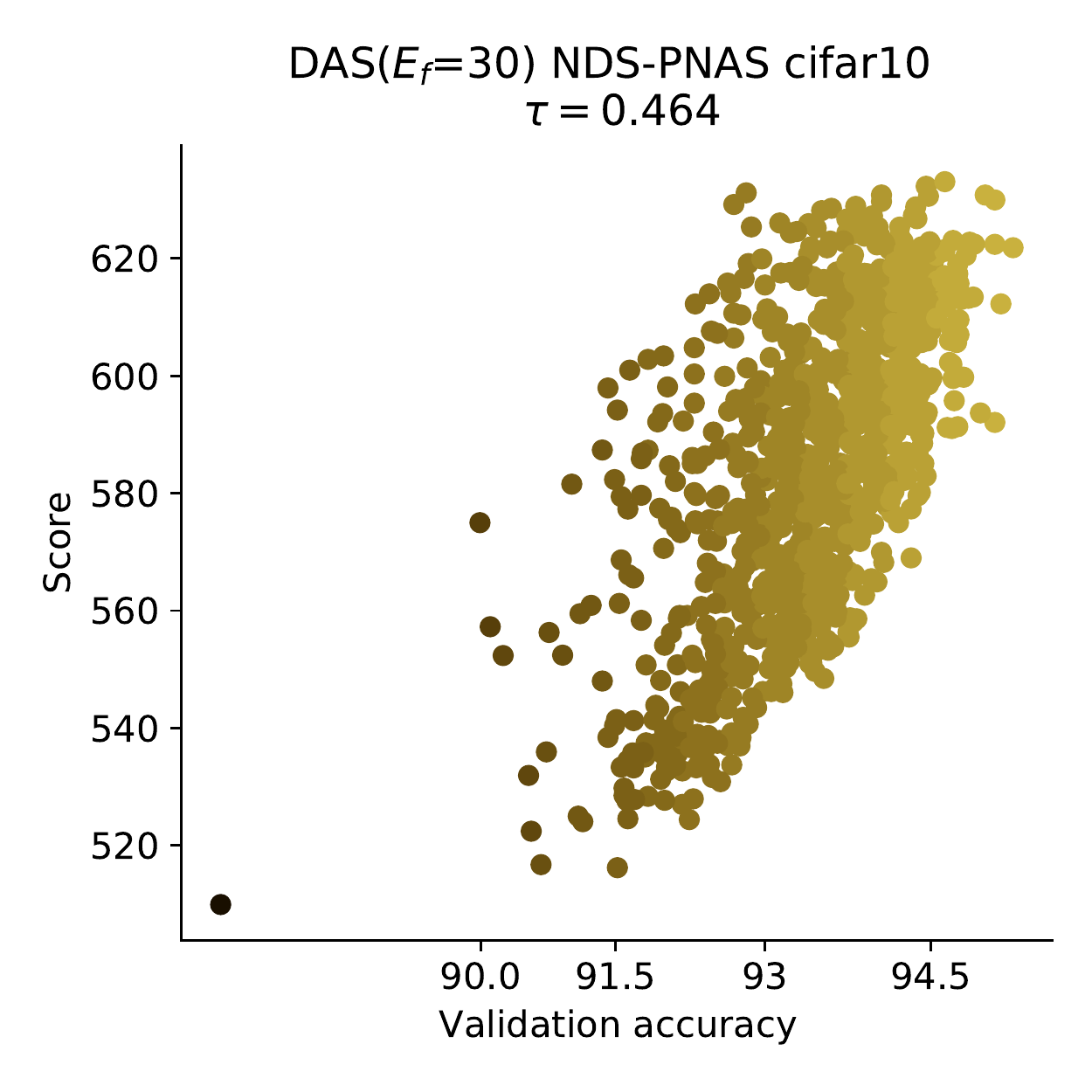}
         \caption{}
         \label{fig:pnas}
     \end{subfigure}
     
    \caption{Plots (a)-(h) visualize the correlation between scores and the classification accuracy on the validation dataset. We show the experimental results from four search spaces NDS-DARTS, NDS-Amoeba, NDS-NASNet, and NDS-PNAS in four colors. 1000 randomly sampled architectures are shown in each plot, and the KTau $\tau$ is reported. The optimal case is that scores and final accuracy are positively correlated, \textit{i.e.}, $\tau$ = 1, and the slope of the line between any two points sampled should be positive.}
    \label{fig:scoreplots} 

\end{figure*}

\begin{table*}
	\centering
        \small
		\vspace{-5pt}
	\caption{Mean and std accuracy of searched architectures on three search spaces. Each method runs 500 times independently, except for the last two lines of the method which runs 100 times. Random is selecting one architecture randomly. The results of DAS for different settings of $E_f$ are reported. In the line \texttt{N}$\geq$1000, \texttt{N} for NAS-Bench-101 and NDS-Amoeba is 1000 and NDS-NASNet is 1200.} 
	\label{tab:search_result}

        \begin{tabular}{c|cc|cc|cc}
            \bottomrule
            & \multicolumn{2}{|c}{NAS-Bench-101} & \multicolumn{2}{|c}{NDS-NASNet} & \multicolumn{2}{|c}{NDS-Amoeba} \\
            Method & Search Cost (s) & Accuracy (\%) & Search Cost (s) & Accuracy (\%)& Search Cost (s) & Accuracy (\%) \\ \hline \hline
            Random & N/A & 90.39 $\pm$ 4.39 & N/A & 91.75 $\pm$ 6.57 & N/A & 92.28 $\pm$ 6.05 \\
            NASWOT~\cite{mellor2021neural}(\texttt{N}=100) &  29.8 & 92.13 $\pm$ 3.98 & 54.6 & 93.35 $\pm$ 1.32 & 51.6 & 92.88 $\pm$ 4.51 \\
            DAS(\texttt{N}=100, $E_f$=0) & 27.8 & 92.11 $\pm$ 3.98 & 49.6 & 93.37 $\pm$ 1.37 & 49.3 & 93.05 $\pm$ 2.62 \\
            NASWOT~\cite{mellor2021neural}(\texttt{N} $\geq$1000) & 167.7 & 92.37 $\pm$ 3.57 & 506.1 & 93.61 $\pm$ 1.08 & 663.9 & 93.04 $\pm$ 0.55 \\
            DAS(\texttt{N}=100, $E_f$=30) & 159.5 & 92.59 $\pm$ 2.67 & 496.1 & 93.77 $\pm$ 0.87 & 607.1 & 93.42 $\pm$ 1.07 \\
            \bottomrule
        \end{tabular}

\end{table*}
\section{Experiment}
\label{sec:experiment}
In this section, we first introduce our proposed dataset DTB and then verify the proposed method in three aspects. Firstly, we compare the methods by the evaluation accuracy over the entire dataset. Secondly, we apply our method to NAS and compare the final accuracy of the searched architectures. Finally, the ablation study of the two improvements proposed in the paper is conducted.

\subsection{Darts-Training-Bench Dataset}
\label{sec:dataset_DTB}
The intention of proposing DTB is to build a dataset containing the training state in the early epochs of the architecture.
The weights state of each architecture in the early training epochs are saved, and we can restore the network in the training process.
To build this dataset, we randomly sampled 500 architectures in the DARTS search space and performed the complete training on CIFAR-10 according to the training settings in the original DARTS experiments. 

Compared to the NAS-Bench-101, the DARTS search space is more carefully designed, and the larger model leads to higher classification accuracy. Compared to the NDS search space, since we follow the training setup of the original paper, the classification accuracy of the trained network is also higher and is competitive with the accuracy of the architecture searched by state-of-the-art NAS methods, such as $\beta$-DARTS~\cite{ye2022b} and DOTS~\cite{gu2021dots}. Considering the consistency of the search space, we quantitatively compared the final accuracy of architecture in DTB and NDS-DARTS. The maximum final accuracy is 97.46\% in DTB while 95.06\% in NDS-DARTS. As for the average value of the final accuracy of the architectures, it is 96.78\% in DTB and 92.65\% in NDS-DARTS. The more accurate and higher final accuracy as the label of the architecture allows for the more accurate evaluation of NAS algorithms.

\subsection{Comparison on Evaluation Accuracy}
\label{sec:exp_evaluation_acc}
We first conduct experiments of evaluation accuracy on different search spaces, and the results are shown in Figure~\ref{fig:compare_nds}. Compared with the WOT score and some well-performed gradient-based metrics, \textit{i.e.}, grad\_norm, and synflow, the method proposed in this paper has significant improvements in all these search spaces. 

Secondly, we visualize the correlation between proxy metrics (WOT and DAS scores) and the validation accuracy of architectures. The visualization in Figure~\ref{fig:scoreplots} shows the correlation in four search spaces. In each search space, the results of WOT score and DAS are reported, and $E_f$ is set to 30. We calculate KTau for each plot and it is clear that KTau has an increasing trend in each dataset, which means that there is a stronger correlation between the proxy scores and validation accuracy. Specifically, for NDS-DARTS, NDS-Amoeba, NDS-NASNet, and NDS-PNAS, KTau improved by a total of 0.076, 0.12, 0.08, and 0.088, respectively. As for search spaces that WOT score is not good at, such as NDS-Amoeba, our method can have nearly double the improvement.

From the visualized scatter plot, the ideal distribution is that the points line up in a diagonally rising line, \textit{i.e.}, scores and validation accuracy are positively correlated. However, for the WOT score, the scatters are almost all aligned in a vertical rectangle. This means that architectures with validation accuracy in the same interval will get very different wot scores, so this method is not effective in separating them. In addition, the evaluation of outliers with poor validation accuracy is inaccurate. This can lead to incorrect selection of these architectures with lower validation accuracy in the case of less sampled architectures. As for our method, the scatter distribution is much closer to the ideal case. The main part is more similar to the diagonal shape rather than a nearly vertical line, and the outliers with lower validation accuracy are in the lower left corner of the plot, allowing for a more accurate search.

\subsection{Comparison on Searched Architecture}
\label{sec:exp_search}
Following the tradition in NAS, we have conducted the search experiments on different search spaces. To make a fair comparison, we adopt the NASWOT~\cite{mellor2021neural} search method, \textit{i.e.}, we randomly sample \texttt{N} architectures, then score these architectures with proxy metrics, and finally select the one with the highest score as the searched architecture. The search results are shown in Table~\ref{tab:search_result}, where the search cost and the mean and std accuracy for a single search are reported.

Table~\ref{tab:search_result} shows the search results on three search spaces. We report the results of DAS with and without the fast training strategy separately. Correspondingly, by adjusting the size of \texttt{N}, we obtain the results for WOT score at the same search cost. Without the fast training strategy, DAS can achieve competitive accuracy at a lower search cost. When using the fast training strategy, DAS can achieve the best in all three search spaces. Although NASWOT searches 10x more architectures than DAS, its performance is still inferior, implying that the quality of the evaluation method is more important than the quantity in the search process, especially when the evaluation is less accurate.

\begin{table*}
	\centering
	\caption{Ablation study of the proposed two strategies on several search spaces. D denotes `Decoupled' and F denotes `Fast training'. $E_f$ is set to 30, and the computational cost is tolerable.} 
	\label{tab:ablation_study}

        \begin{tabular}{cc|ccccccc}
            \bottomrule
            D & F &  NAS-Bench-101 & NDS-PNAS & NDS-ENAS & NDS-DARTS & NDS-Ameoba & NDS-NASNet & DTB      \\ \hline \hline
            & & 0.2803 & 0.3639 & 0.3882 & 0.4807 & 0.2092 & 0.2997 & 0.3828 \\
            \checkmark & & 0.2851 & 0.3690 & 0.3901 & 0.4850 & 0.2106 & 0.3167 & 0.3875 \\
            & \checkmark & 0.3167 & 0.4147 & 0.4277 & 0.5319 & 0.2856 & 0.3407 & 0.3899 \\
            \checkmark & \checkmark & \textbf{0.3291} & \textbf{0.4437} & \textbf{0.4487} & \textbf{0.5576} & \textbf{0.3269} & \textbf{0.3798} & \textbf{0.3992} \\ \bottomrule
        \end{tabular}

\end{table*}

\begin{table}
	\centering
        \small
	\caption{Computational cost (seconds) on fast training. The multiplicative relationship is calculated based on the cost without using fast training.} 
	\label{tab:computational}

        \begin{tabular}{c|ccc}
            \bottomrule
            Epoch & NDS-ENAS & NDS-Ameoba & NDS-NASNet    \\ \hline \hline
            0 & 2322(1.0x) & 3368(1.0x) & 2635(1.0x) \\
            10 & 11081(4.8x) & 17766(5.3x) & 12560(4.8x) \\ 
            30 & 23382(10.0x) & 39413(11.7x) & 27721(10.5x) \\ \bottomrule
        \end{tabular}

\end{table}

\subsection{Ablation Study}
\label{sec:ablation}
To verify the effectiveness of the two strategies in the proposed DAS, exhaustive ablation experiments are performed on multiple search spaces. The KTau values of experiments are shown in Table~\ref{tab:ablation_study}. For convenient comparison, the best results on each search space are bolded. It is clear that the highest KTau values can be achieved when decoupled WOT and fast training are used together. When neither of the two strategies proposed in this paper is used, \textit{i.e.}, when the original WOT score is used, only the lowest KTau value is available on each search space.

\begin{figure}
    \centering
    \includegraphics[width=1.0\linewidth]{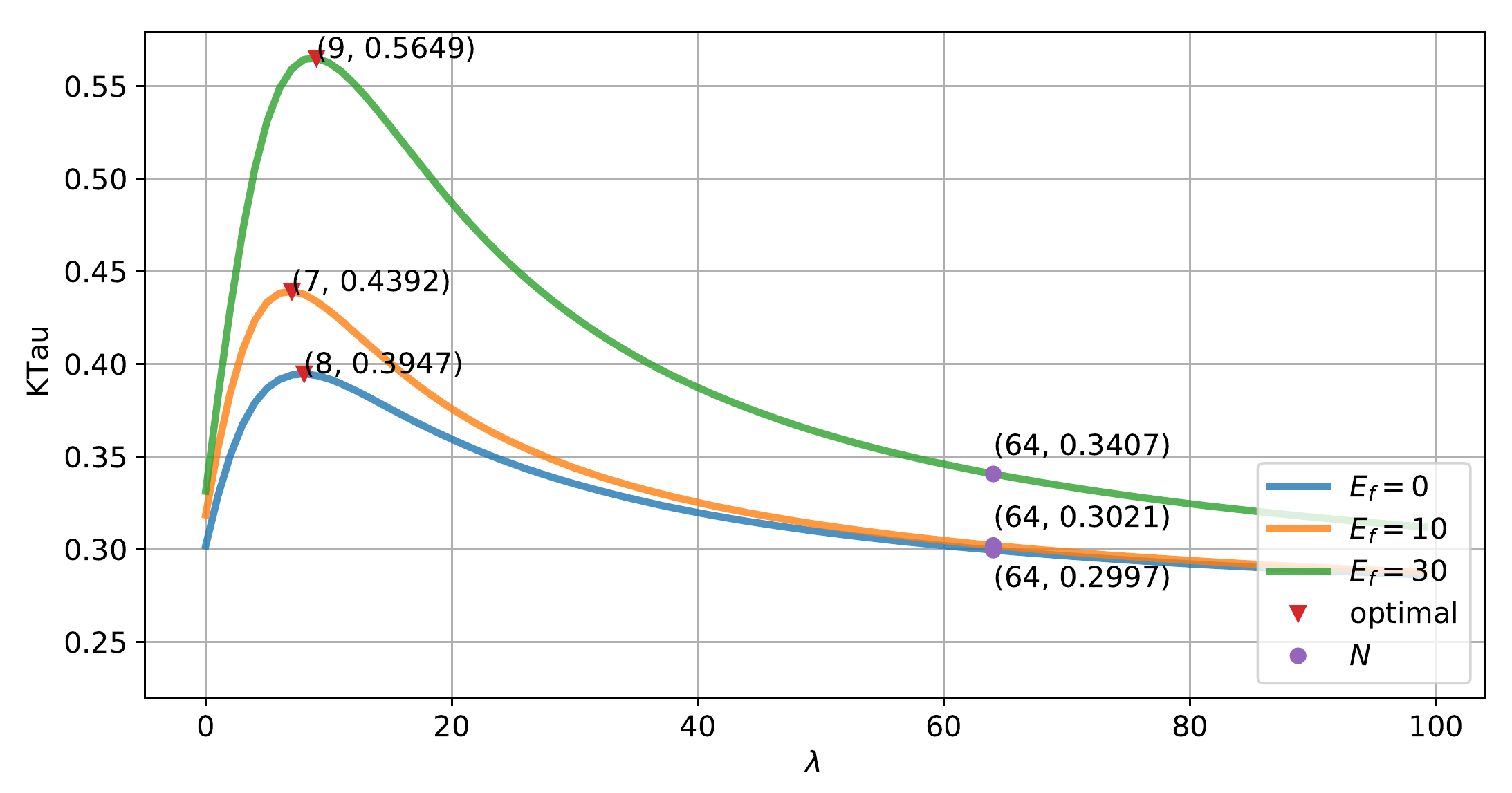}
    \vspace{-15pt}
    \caption{Decoupled WOT after fast training. The search space is NDS-NASNet.}
    \vspace{-5pt}
    \label{fig:fast_training_ablation}
\end{figure}

When only one strategy is used, the KTau values are all increased to varying degrees. Although the improvement is smaller when decoupled WOT is used, the decoupled strategy simply gets a more accurate evaluation by reorganizing the atomic metrics, which does not impose an additional computational cost. When only the fast training strategy is used, there is a more significant increase in KTau values. This is reasonable because there is an increased cost in computational demand. Also, it is notable that using the decoupled WOT strategy after the fast training strategy gives a bigger boost. Taking NDS-NASNet as an example, the result of searching for the optimal $\lambda$ after fast training is shown in Figure~\ref{fig:fast_training_ablation}. Most of the search spaces showed this phenomenon, which means that when the two strategies are used together, they can have a greater improvement.

Finally, an ablation study about the additional computational cost of fast training strategy has been conducted, and the results are shown in Table~\ref{tab:computational}. The time required for training is proportional to the size of the architecture. Although our method requires about ten times more computational cost than the original WOT, the extra computational cost is acceptable considering that the WOT score itself is a very low-cost method. With almost 5,000 architectures in each NDS dataset, the time required to evaluate an architecture individually is within 10 seconds as calculated in Table~\ref{tab:computational}, which is a very significant improvement on the evaluation method of the original paper~\cite{pham2018efficient,real2019regularized,zoph2018learning}.

\section{Discussion}
\label{sec:discussion}
In this section, we will discuss two aspects. The first is how to determine the optimal combination approach and the optimal coefficients. The second is which atomic metric should be used to evaluate the architecture.

In this paper, although we simply evaluate the architecture by taking the logarithm of the atomic metrics and summing them, it is also very difficult to find how to set the optimal coefficients, \textit{i.e.}, $\lambda$, in this combinatorial form. The optimal $\lambda$ is influenced by various factors, including the search space, the training dataset, and the training settings. Our proposed DTB dataset and NDS-DARTS dataset both contain architectures sampled in the DARTS space, but the optimal $\lambda$ differs due to the different training settings (as seen in Figure~\ref{fig:de_wot}). We have tried to find associations with the optimal $\lambda$ from these factors and have tried to find a way to adaptively give the optimal $\lambda$ settings for different datasets, but is unsuccessful.

Furthermore, we have also tried to evaluate the architecture using other atomic metrics together. For example, the number of activation layers with optimal coefficient settings can greatly improve the accuracy of the evaluation. However, we also face the first issue: how to find the optimal coefficients adaptively? Experimental results on multiple datasets show that the optimal coefficients are variable, so we cannot utilize the number of activation layers as an atomic metric in the evaluation.

\section{Conclusion}
\label{sec:conclusion}
The promising proxy evaluation metric WOT score in NAS can complete the evaluation of the architecture with little computational cost, dramatically accelerating the evaluation process in NAS. This paper notes that WOT is not an atomic metric, and decouples it into two atomic metrics. Based on this, we propose DAS to accurately evaluate the architecture. In addition, the fast training strategy is also able to significantly improve the evaluation accuracy of our method. Moreover, we propose a dataset DTB containing the accuracy of the complete training of the architecture and the status of the architecture in early training epochs. Experimental results on several datasets show a significant improvement. The work in this paper also has some limitations. How to set the metric coefficient adaptively and what kind of atomic metrics should be used are our future work.

{\small
\bibliographystyle{ieee_fullname}
\bibliography{PaperForReview}

\begin{thebibliography}{10}\itemsep=-1pt

\bibitem{abdelfattah2020zero}
Mohamed~S Abdelfattah, Abhinav Mehrotra, {\L}ukasz Dudziak, and Nicholas~Donald
  Lane.
\newblock Zero-cost proxies for lightweight nas.
\newblock In {\em {Proc. ICLR}}, 2020.

\bibitem{deng2017peephole}
Boyang Deng, Junjie Yan, and Dahua Lin.
\newblock Peephole: predicting network performance before training.
\newblock {\em arXiv preprint arXiv:1712.03351}, 2017.

\bibitem{devries2017improved}
Terrance DeVries and Graham~W Taylor.
\newblock Improved regularization of convolutional neural networks with cutout.
\newblock {\em arXiv preprint arXiv:1708.04552}, 2017.

\bibitem{elsken2019neural}
Thomas Elsken, Jan~Hendrik Metzen, and Frank Hutter.
\newblock Neural architecture search: A survey.
\newblock {\em The Journal of Machine Learning Research}, 20(1):1997--2017,
  2019.

\bibitem{floreano2008neuroevolution}
Dario Floreano, Peter D{\"u}rr, and Claudio Mattiussi.
\newblock Neuroevolution: from architectures to learning.
\newblock {\em Evolutionary Intelligence}, 1(1):47--62, 2008.

\bibitem{ghiasi2019fpn}
Golnaz Ghiasi, Tsung-Yi Lin, and Quoc~V Le.
\newblock Nas-fpn: learning scalable feature pyramid architecture for object
  detection.
\newblock In {\em {Proc. CVPR}}, pages 7036--7045, 2019.

\bibitem{gu2021dots}
Yu-Chao Gu, Li-Juan Wang, Yun Liu, Yi Yang, Yu-Huan Wu, Shao-Ping Lu, and
  Ming-Ming Cheng.
\newblock Dots: decoupling operation and topology in differentiable
  architecture search.
\newblock In {\em {Proc. CVPR}}, pages 12311--12320, 2021.

\bibitem{he2016deep}
Kaiming He, Xiangyu Zhang, Shaoqing Ren, and Jian Sun.
\newblock Deep residual learning for image recognition.
\newblock In {\em {Proc. CVPR}}, pages 770--778, 2016.

\bibitem{krizhevsky2009learning}
Alex Krizhevsky, Geoffrey Hinton, et~al.
\newblock Learning multiple layers of features from tiny images.
\newblock 2009.

\bibitem{lee2018snip}
Namhoon Lee, Thalaiyasingam Ajanthan, and Philip Torr.
\newblock Snip: single-shot network pruning based on connection sensitivity.
\newblock In {\em {Proc. ICLR}}, 2018.

\bibitem{liu2018progressive}
Chenxi Liu, Barret Zoph, Maxim Neumann, Jonathon Shlens, Wei Hua, Li-Jia Li, Li
  Fei-Fei, Alan Yuille, Jonathan Huang, and Kevin Murphy.
\newblock Progressive neural architecture search.
\newblock In {\em {Proc. ECCV}}, pages 19--34, 2018.

\bibitem{liu2019darts}
Hanxiao Liu, Karen Simonyan, and Yiming Yang.
\newblock {DARTS:} differentiable architecture search.
\newblock In {\em {Proc. ICLR}}, 2019.

\bibitem{liu2021survey}
Yuqiao Liu, Yanan Sun, Bing Xue, Mengjie Zhang, Gary~G Yen, and Kay~Chen Tan.
\newblock A survey on evolutionary neural architecture search.
\newblock {\em IEEE Trans. on Neural Networks and Learning Systems}, 2021.

\bibitem{Liu_2021_ICCV}
Yuqiao Liu, Yehui Tang, and Yanan Sun.
\newblock Homogeneous architecture augmentation for neural predictor.
\newblock In {\em {Proc. ICCV}}, pages 12249--12258, 2021.

\bibitem{mellor2021neural}
Joe Mellor, Jack Turner, Amos Storkey, and Elliot~J Crowley.
\newblock Neural architecture search without training.
\newblock In {\em {Proc. ICML}}, pages 7588--7598, 2021.

\bibitem{pham2018efficient}
Hieu Pham, Melody Guan, Barret Zoph, Quoc Le, and Jeff Dean.
\newblock Efficient neural architecture search via parameters sharing.
\newblock In {\em {Proc. ICML}}, pages 4095--4104. PMLR, 2018.

\bibitem{radosavovic2019network}
Ilija Radosavovic, Justin Johnson, Saining Xie, Wan-Yen Lo, and Piotr
  Doll{\'a}r.
\newblock On network design spaces for visual recognition.
\newblock In {\em {Proc. ICCV}}, pages 1882--1890, 2019.

\bibitem{real2019regularized}
Esteban Real, Alok Aggarwal, Yanping Huang, and Quoc~V Le.
\newblock Regularized evolution for image classifier architecture search.
\newblock In {\em {Proc. AAAI}}, volume~33, pages 4780--4789, 2019.

\bibitem{real2017large}
Esteban Real, Sherry Moore, Andrew Selle, Saurabh Saxena, Yutaka~Leon Suematsu,
  Jie Tan, Quoc~V Le, and Alexey Kurakin.
\newblock Large-scale evolution of image classifiers.
\newblock In {\em {Proc. ICML}}, pages 2902--2911, 2017.

\bibitem{saxena2016convolutional}
Shreyas Saxena and Jakob Verbeek.
\newblock Convolutional neural fabrics.
\newblock {\em {Proc. NeurIPS}}, 29:4053--4061, 2016.

\bibitem{simonyan2014very}
Karen Simonyan and Andrew Zisserman.
\newblock Very deep convolutional networks for large-scale image recognition.
\newblock {\em arXiv preprint arXiv:1409.1556}, 2014.

\bibitem{tanaka2020pruning}
Hidenori Tanaka, Daniel Kunin, Daniel~L Yamins, and Surya Ganguli.
\newblock Pruning neural networks without any data by iteratively conserving
  synaptic flow.
\newblock {\em {Proc. NeurIPS}}, 33:6377--6389, 2020.

\bibitem{Wang2020Picking}
Chaoqi Wang, Guodong Zhang, and Roger Grosse.
\newblock Picking winning tickets before training by preserving gradient flow.
\newblock In {\em {Proc. ICLR}}, 2020.

\bibitem{wen2020neural}
Wei Wen, Hanxiao Liu, Yiran Chen, Hai Li, Gabriel Bender, and Pieter-Jan
  Kindermans.
\newblock Neural predictor for neural architecture search.
\newblock In {\em {Proc. ECCV}}, pages 660--676, 2020.

\bibitem{ye2022b}
Peng Ye, Baopu Li, Yikang Li, Tao Chen, Jiayuan Fan, and Wanli Ouyang.
\newblock B-darts: beta-decay regularization for differentiable architecture
  search.
\newblock In {\em {Proc. CVPR}}, pages 10874--10883, 2022.

\bibitem{ying2019bench}
Chris Ying, Aaron Klein, Eric Christiansen, Esteban Real, Kevin Murphy, and
  Frank Hutter.
\newblock Nas-bench-101: towards reproducible neural architecture search.
\newblock In {\em {Proc. ICML}}, pages 7105--7114, 2019.

\bibitem{zoph2016neural}
Barret Zoph and Quoc~V Le.
\newblock Neural architecture search with reinforcement learning.
\newblock {\em arXiv preprint arXiv:1611.01578}, 2016.

\bibitem{zoph2018learning}
Barret Zoph, Vijay Vasudevan, Jonathon Shlens, and Quoc~V Le.
\newblock Learning transferable architectures for scalable image recognition.
\newblock In {\em {Proc. CVPR}}, pages 8697--8710, 2018.

\end{thebibliography}
}

\end{document}